\documentclass{article}
\pdfoutput=1

\usepackage{amsmath}

\usepackage{arxiv}
\usepackage{graphicx}
\usepackage{siunitx,etoolbox}
\usepackage{longtable}
\usepackage{import}

\usepackage[utf8]{inputenc} % allow utf-8 input
\usepackage[T1]{fontenc}    % use 8-bit T1 fonts
\usepackage{hyperref}       % hyperlinks
\usepackage{url}            % simple URL typesetting
\usepackage{booktabs}       % professional-quality tables
\usepackage{amsfonts}       % blackboard math symbols
\usepackage{nicefrac}       % compact symbols for 1/2, etc.
\usepackage{microtype}      % microtypography

\usepackage{lipsum}
\usepackage[square,sort,comma,numbers]{natbib}

\newcommand{\vb}{\mathbf{v}}
\newcommand{\xb}{\mathbf{x}}
\newcommand{\yb}{\mathbf{y}}
\newcommand{\wb}{\mathbf{w}}

\title{BEHRT: Transformer for Electronic Health Records}
\author{
 Yikuan Li \\
  Deep Medicine\\
  Oxford Martin School\\
  University of Oxford\\
  Oxford, UK \\
  \texttt{yikuan.li@georgeinstitute.ox.ac.uk} \\
   \And
 Shishir Rao \thanks{Shishir Rao and Yikuan Li have equally contributed to this work as first authors. Shishir Rao is Corresponding Author for this work.} \\
  Deep Medicine\\
  Oxford Martin School\\
  University of Oxford\\
  Oxford, UK \\
  \texttt{shishir.rao@georgeinstitute.ox.ac.uk} \\
   \AND
   Jose Roberto Ayala Solares \\
   Deep Medicine\\
   Oxford Martin School \\
   University of Oxford\\
   \And
  Abdelaali Hassaine\\
   Deep Medicine\\
   Oxford Martin School \\
   University of Oxford\\
   \And
   Dexter Canoy \\
   Deep Medicine\\
   Oxford Martin School \\
   University of Oxford\\
    \And
   Yajie Zhu\\
   Deep Medicine\\
   Oxford Martin School \\
   University of Oxford\\
      \And
   Kazem Rahimi\\
   Deep Medicine\\
   Oxford Martin School \\
   University of Oxford\\
    \And
    Gholamreza Salimi-Khorshidi \\
   Deep Medicine\\
   Oxford Martin School \\
   University of Oxford\\
}

\begin{document}

\maketitle

\begin{abstract}
Today, despite decades of developments in medicine and the growing interest in precision healthcare, vast majority of diagnoses happen once patients begin to show noticeable signs of illness. Early indication and detection of diseases, however, can provide patients and carers with the chance of early intervention, better disease management, and efficient allocation of healthcare resources. The latest developments in machine learning (more specifically, deep learning) provides a great opportunity to address this unmet need. In this study, we introduce BEHRT: A deep neural sequence transduction model for EHR (electronic health records), capable of multitask prediction and disease trajectory mapping. When trained and evaluated on the data from nearly 1.6 million individuals, BEHRT shows a striking absolute improvement of 8.0-10.8\%, in terms of Average Precision Score, compared to the existing state-of-the-art deep EHR models (in terms of average precision, when predicting for the onset of 301 conditions). In addition to its superior prediction power, BEHRT provides a personalised view of disease trajectories through its attention mechanism; its flexible architecture enables it to incorporate multiple heterogeneous concepts (e.g., diagnosis, medication, measurements, and more) to improve the accuracy of its predictions; and its (pre-)training results in disease and patient representations that can help us get a step closer to interpretable predictions.
\end{abstract}

%\keywords{machine learning \and deep learning \and Transformer \and BERT \and medical records \and EHR \and EMR \and Natural language processing \and Prediction}

\section{Introduction}
\label{intro}

The field of precision healthcare aims to improve the provision of care through precise and personalised prediction, prevention, and intervention.

In recent years, advances in deep learning (DL) - a subfield of machine learning (ML) - has led to great progress towards personalised predictions in cardiovascular medicine, radiology, neurology, dermatology, ophthalmology, and pathology, just to name a few. For instance,~\cite{Ardila2019} introduced a DL model that can predict the risk of lung cancer from a patient's tomography images with a striking 94.4\% accuracy;~\cite{Poplin2018} showed that DL can predict a range of cardiovascular risk factors from just a retinal fundus photograph; and the list continues (more examples can be found in~\cite{Topol2019} and~\cite{Esteva2019AHealthcare.}). A key contributing factor to this success, in addition to the developments in DL algorithms, was the massive influx of large multimodal biomedical data, including but not limited to, mega cohorts such as UK Biobank~\cite{Sudlow2015UKAge}, and routinely-collected health data such as electronic health records (EHR)~\cite{Shickel2018DeepAnalysis}.

In the recent years, the adoption of EHR systems has greatly increased; percent of hospitals in the US and UK that have adopted EHR systems now exceeds 84\% and 94\%, respectively~\cite{Meeting2018, Parasrampuria2019Hospitals2015-2017}. As a result, EHR systems of a national (and/or a large) medical organisation now are likely to capture data from millions of individuals over many years (or decades, sometimes). Each individual's EHR can link data from many sources (e.g., doctor visits and hospital episodes) and hence contain ``concepts'' such as diagnoses, interventions, lab tests, clinical narratives, and more. Each instance of a concept can mean a single or multiple data points; just a single hospitalisation alone, for instance, can generate thousands of data points for an individual, whereas a diagnosis can be a single data point (i.e., an ICD code). This makes large-scale EHR a uniquely rich source of insight and an unrivalled data for training data-hungry ML models.

In traditional research on EHR (including the ones using ML), individuals are represented to models as a vector of attributes, or "features" ~\cite{Rahimian2018PredictingRecords}. This approach relies on experts' ability to define the appropriate features, and design the model's structure (i.e., answering questions such as ``what are the key features for this prediction?'' or, ``which features should have interactions with one another?''). Recent developments in deep learning, however, provided us with models that can learn useful representations (e.g., of individuals, concepts, or an entire record) from raw or minimally-processed data, with minimal need for expert guidance. This happens through a sequence of layers, each employing a large number of simple linear and nonlinear operations to map their corresponding inputs to a representation; the progress from layer to layer, is expected to result in a final representation in which the data points form distinguishable patterns.

As one of the earliest works on applying deep learning to EHR, Liang et al~\citep{Liang2014} showed that deep neural networks can outperform SVM and decision tree paired with manual feature engineering, over a number prediction tasks on a number of different datasets. In another early work in this space, Tran et al~\cite{Tran2015LearningeNRBM} proposed the use of restricted Boltzmann machines (RBM) for learning a distributed representation of EHR, which was shown to outperform the manual feature extraction, when predicting the risk of suicide from individuals' EHR. In a similar approach, Miotto et al~\cite{Miotto2016DeepRecordsb} employed a stack of denoising autoencoders (SDA) instead of RBM, and showed that it outperforms many popular feature extraction and feature transformation approaches (e.g., PCA, ICA and Gaussian mixture models) for providing classifiers with useful features to predict the onset of a number of diseases from EHR.

These early works on the application of DL to EHR did not take into account the subtleties of EHR data (e.g., the irregularity of the inter-visit intervals, and the temporal order or events, to name a few). In attempt to address this, Nguyen et al~\cite{Nguyen2017} introduced a convolutional neural network (CNN) model called Deepr (Deep record) for predicting the probability of readmission; they treated one's medical history as a sequence of concepts (e.g., diagnosis and medication) and inserted a special word between each pair of consecutive visits to denote the time difference between them. In another similar attempt, Choi et al.~\cite{Choi2016DoctorNetworks.} introduced a shallow recurrent neural network (RNN) model to predict the diagnoses and medications that are likely to occur in the subsequent visit. Both these works employed some form of embedding to map the non-numeric medical concepts to an algebraic space in which the sequence models can operate.

One of the improvements that was next introduced to the DL models of EHR aimed to enable them to capture the long-term dependencies among events (e.g., key diagnoses such as diabetes can stay a risk factor over a person's life, even decades after their first occurrence; certain surgeries may prohibit certain future interventions). Pham et al~\cite{Pham2016} introduced a Long Short-Term Memory (LSTM) architecture with attention mechanism, called DeepCare, which outperformed standard ML techniques, LSTM, and plain RNN in tasks such as prediction of the onset of diabetes. In a similar development, Choi et al~\cite{Choi2016a} proposed a model based on reverse-time attention mechanism to consolidate past influential visits using an end-to-end RNN model named RETAIN for the prediction of heart failure. RETAIN outperformed most of the models at the time of its publication and provided a decent baseline for the medical deep learning research~\cite{Solares}.

In this study, given the success of deep sequence models and attention mechanisms in the past DL research for EHR, we aim to build on some of the latest developments in deep learning and natural language processing (NLP) -- more specifically, Transformer architecture~\cite{Devlin2018} -- while taking into account various EHR-specific challenges, and provide improved accuracy for the prediction of future diagnoses. We named our model BEHRT (i.e., BERT for EHR), due to architectural similarities that it has with (and our original inspirations that came from) BERT~\cite{Devlin2018}; one of the most powerful Transformer-based architectures in NLP.

\section{Materials and Methods}
\label{mm}
In this section, after providing an introduction to our EHR data, we will introduce the earlier works that inspired BEHRT, as well as the novel features that this new architecture contributes to the field.

\subsection{Data}
\label{cprd}
In this study, we used CPRD (Clinical Practice Research Datalink)~\cite{Herrett2015, Solares}; it contains longitudinal primary care data from a network of 674 general physician (GP) practices in the UK, which is linked to secondary care (i.e., hospital episode statistics, or HES) and other health and administrative databases (e.g., office for national statistics' death registration). Around 1 in 10 GP practices (and nearly 7\% of the population) in the UK contribute data to CPRD; it covers 35 million patients, among whom nearly 10 million are currently registered patients~\cite{Herrett2015}.
CPRD is broadly representative of the population by age, sex, and ethnicity. It has been extensively validated and is considered as the most comprehensive longitudinal primary care database~\cite{WALLEY19971097}, with several large-scale epidemiological reports~\cite{Emdin2015UsualAdults,Emdin2017,Herrett2015} adding to its credibility. % [RK] please correct this citations in this paragraph that I picked from Ali's paper/bib file

HES, on the other hand, contains data on hospitalisations, outpatient visits, accident and emergency for all admissions to National Health Service (NHS) hospitals in England~\cite{Lee2002}. Approximately 75\% of the CPRD GP practices in England (58\% of all UK CPRD GP practices) participate in patient-level record linkage with HES, which is performed by the Health and Social Care Information Centre~\cite{Mohseni2017}. In this study, we only considered the data from GP practices that consented to (and hence have) record linkage with HES. The importance of primary care at the centre of the national health system in the UK, the additional linkages, and all the aforementioned properties, make CPRD one of the most suitable EHR datasets in the world for data-driven clinical/medical discovery and machine learning.  % [RK] please correct this citations in this paragraph that I picked from Ali's paper/bib file

\subsection{Preprocessing of CPRD}
\label{preproc}
\begin{figure}[h]
\centering
\includegraphics[width=1\textwidth]{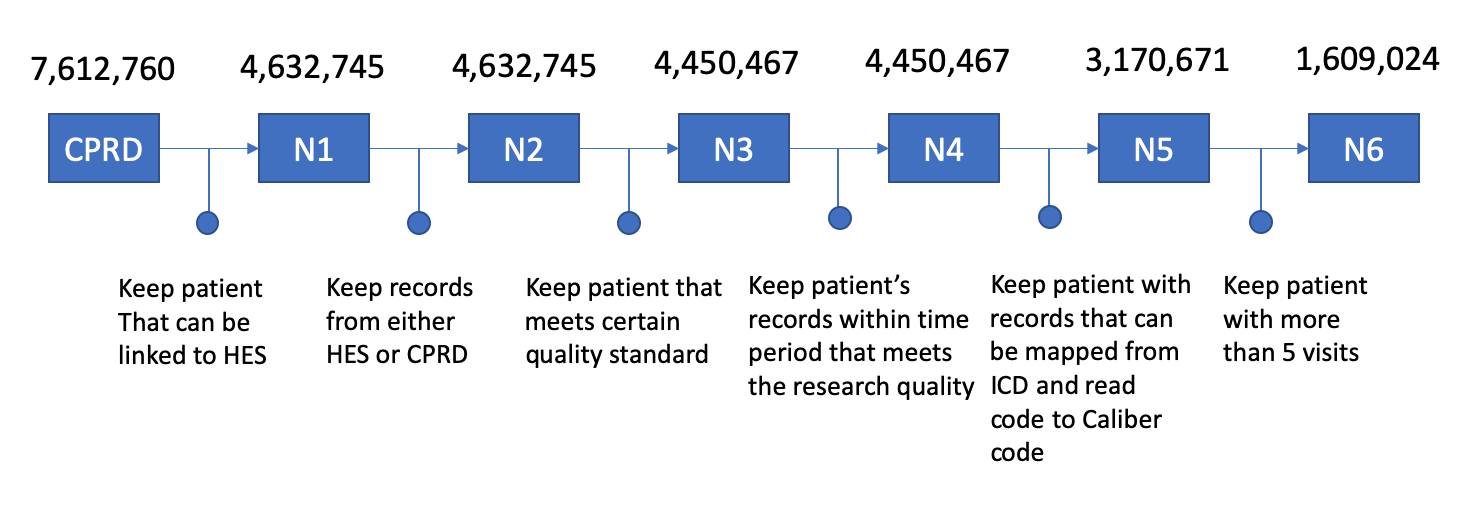}
	\caption{Linkage and filtering of CPRD data. This flow lists all the key steps of our data cleaning and linkage procedure. At each step, the number of patients that are included is shown.
	% [RK] do you also check if the record is "up to standard"? Also, show the precise number in the flowchart (and use comma to make them easy to read). Also, when you say "map ICD ... to caliber", it is not clear what is the criteria of keeping a record after that; please make it clearer like the "keep *" statements you use in the rest of the fig. I really like what Ali did in his paper; can we have a similar one? Note that, your use of CPRD is one of your paper's strengths and we must somehow show it.
	}
	\label{fig:cprd_incl}
\end{figure}

We start our preprocessing with 8 million patients; we only included patients that are eligible for linkage to HES and meet CPRD's quality standard (i.e., using the flags and time windows that CPRD provides to indicate the quality of one's EHR). Furthermore, to only keep the records that have enough history to be useful for prediction, we only kept individuals who have at least 5 visits in their EHR. At the end of this process, we are left with $P=1.6$ million patients to train and evaluate BEHRT on. More details on the steps we took and the number of patients after each one of them can be seen in Fig~\ref{fig:cprd_incl}.

In CPRD, diseases are classified using Med Code (which can be simply mapped to NHS's standard Read Code~\cite{NHS}) and ICD-10~\cite{WHO} schemes, for primary and hospital care, respectively. In the ICD-10 universe, one can define diseases at the desired level of granularity that is appropriate for the analysis of interest by simply choosing the level of hierarchy they want to operate at; for instance, operating ICD-10 chapter level will lead to 22 diseases, while operation at full ICD-10 code level will lead to 10,000 diseases. With Med Code, however, such a hierarchy does not exist and hence one needs to carry out exhaustive disease review, in order to define the diseases of interest for a given analysis. To alleviate this extra data-processing burden, we first mapped Med Codes to Read Codes (using the mapping provided by CPRD) and excluded the procedure codes. After that, we mapped both ICD-10 codes (at level 4) and Read Codes to Caliber codes~\cite{Kuan2019}, which is an expert checked mapping dictionary from University College London. Eventually, this resulted in a total of $G=301$ codes for diagnoses. We denote the list of all these diseases as $D=\{d_i\}_{i=1}^G$, where $d_i$ denotes the $i$th disease code.

\begin{figure}[h]
    \centering
    \includegraphics[width=1\textwidth]{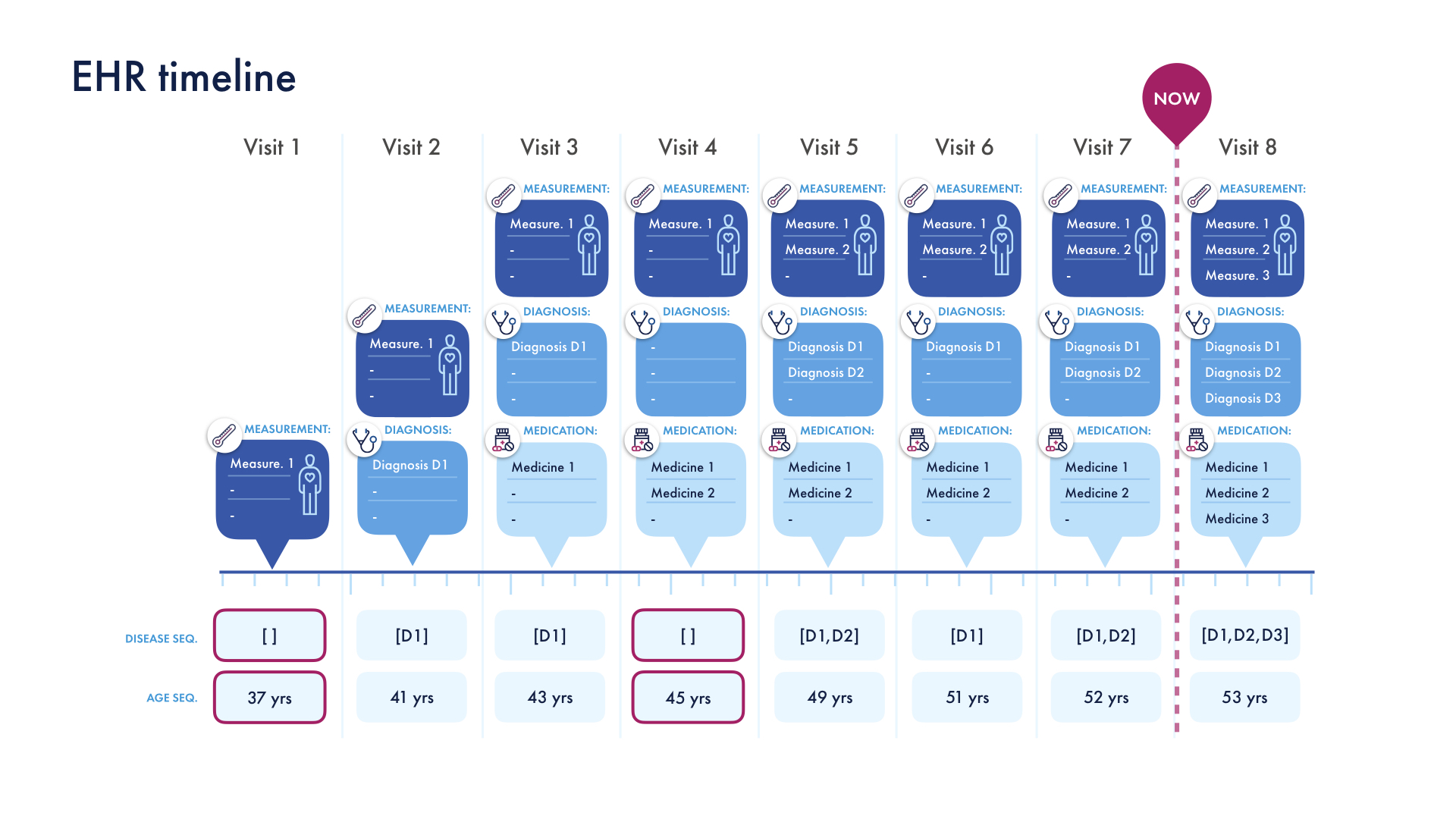}
    \caption{Preparation of CPRD data for BEHRT. An example patient's EHR sequence can be seen in (a), which consists of 8 visits. In each visit, the record can consist of concepts such as diagnoses, medications and measurements; all these values are artificial and for illustration purposes. For this study, we are only interested in age and diagnoses. Therefore, as shown in at the bottom of the figure, we have taken only the diagnosis and age subset of the record to form the necessary sequences. This resulting sequence is what we represent every patient's EHR as to our modelling process. Note that, the visits shown in purple boxes are not going to be represented to the model, due to them lacking diagnoses.}
    \label{fig:cprd_introduction}
\end{figure}

For each patient $p \in \{1, 2, \ldots, P\}$ the medical history consists of a sequence of visits to GP and hospitals; each visit can contain concepts such as diagnosis, medications, measurements and more. In this study, however, we are only considering the diagnoses; we denote each patient's EHR as $V_p=\{\vb_p^1, \vb_p^2, \vb_p^3, \ldots, \vb_p^{n_p}\}$, where $n_p$ denotes the number of visits in patient $p$'s EHR, and $v_p^j$ contains the diagnoses in the $j$th visit, which can be viewed as a list of $m_p^j$ diagnoses (i.e., $\vb_p^j=\{d_1,\ldots, d_{m_p^j}\}$). In order to prepare the data for BEHRT, we order the visits (hence diseases) temporally, and introduce a term to denote the start of medical history (i.e., $CLS$), and the space between visits (i.e., $SEP$), which results in a new sequence, $V_p=\{CLS, \vb_p^1, SEP, \vb_p^2, SEP, \ldots, \vb_p^{n_p}, SEP\}$, that from now on will be how we see/denote each patient's EHR as. This process is illustrated in Figure~\ref{fig:cprd_introduction}.

\subsection{BEHRT: A Transformer-based Model for EHR}
\label{behrt}
In this study, we aim to use a given patient's past EHR to predict his/her future diagnoses (if any), as a multi-task classification problem; this will result in a single predictive model that scales across a range of diseases (as opposed to needing to train one predictive model per disease). Modelling of EHR sequences requires dealing with four key challenges~\cite{Pham2016}: (C.1) complex and nonlinear interactions among past, present and future concepts; (C.2) long-term dependencies among concepts (e.g., diseases occurring early in the history of a patient effecting events far later in future); (C.3) difficulties of representing multiple heterogeneous concepts of variable sizes and forms to the model; and (C.4) the irregular intervals between consecutive visits.

Similarities between sequences in EHR and natural language led to successful use of techniques such as BoW~\cite{Miotto2016DeepRecordsb}, Skip-gram~\cite{Choi2016DoctorNetworks.}, RNN~\cite{Choi2016a}, and attention~\cite{Pham2016,Cho2013} ({\em a la} their NLP usage) for learning complex EHR representations in the past. In this study, we get our inspiration from the striking success that Transformers~\cite{Vaswani2017}, and more specifically, a Transformer-based architecture known as BERT~\cite{Devlin2018}. By depicting diagnoses as words, the content of in each visit as a sentence, and a patient's entire medical history as a document, we facilitate the the use of multi-headed self-attention, positional encoding, and masked language models (MLM), for EHR - under a new architecture we call BEHRT.

\begin{figure}[h]
    \centering
    \includegraphics[width=0.9\textwidth]{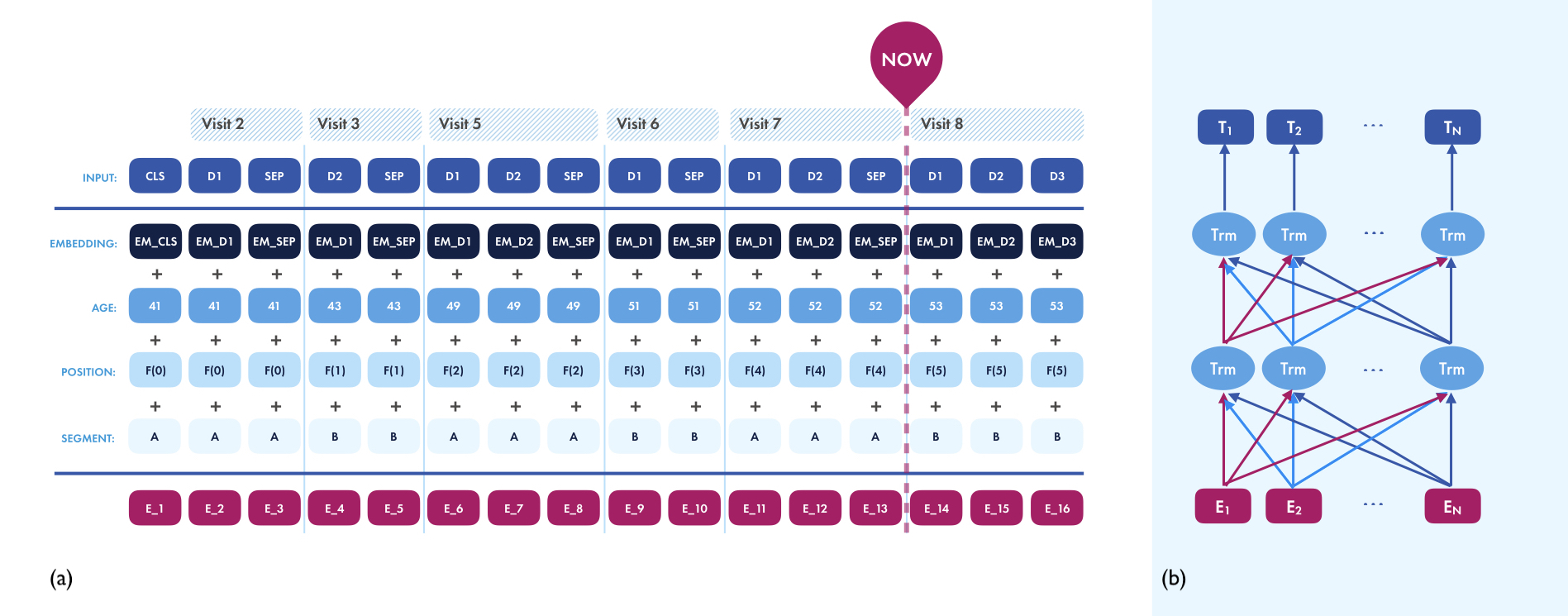}
    \caption{BEHRT architecture. Using the artificial data shown in Figure~\ref{fig:cprd_introduction}, (a) shows how BEHRT sees one's EHR. In addition to diagnosis and age, BEHRT also employs an encoding for event's positions (shown as POSITION) to help it understand each event's place in the sequence, and a visit separator (shown as SEGMENT) to help it understand the start and end of a visit. The combination of all these results is a final embedding shown at the bottom of (a), which will be the latent contextual representation of one's EHR at a given visit. Using this pre-trained embedding, BEHRT's deep Transformer-based model -- shown in (b) -- learns to predict the diseases in one's future.}
    \label{fig:BEHRT}
\end{figure}

We refer readers to the original papers~\cite{Vaswani2017, Devlin2018} for an exhaustive background description for both Transformer and BERT. Figure~\ref{fig:BEHRT}B illustrates BEHRT's architecture, which is designed to pre-train deep bidirectional representations of medical concepts by jointly conditioning on both left and right contexts in all layers. The pre-trained representations can be simply employed for a wide range of downstream tasks, e.g., prediction of the next diseases, and disease phenomapping. Such bidirectional contextual awareness of BEHRT's representations is a big advantage when dealing with EHR, where due to variabilities in practice of care and/or simply due to random events, the order at which certain diseases happen can be reversed, or the time interval between two diagnoses can be shorter or longer than actually recorded.

BEHRT, has many structural advantages over many of the previous methods for modelling EHR data. Firstly, we use feedforward neural networks to model time sensitive EHR data by examining the various forms of sequential order within the data (e.g. age, order of visits) instead of using traditional state-of-the-art RNN and CNN that were explored in the past~\cite{Nguyen2017, Choi2016a}. Recurrent neural networks are known to be notoriously hard to train, due to their exploding and vanishing gradient problems~\cite{Pascanu2012OnNetworks}; these issues hamper these networks' ability to learn (particularly, when dealing with long sequential data). On the other hand, convolutional neural networks only capture limited amount of information with convolutional kernels in the lower layers, and need to expand their receptive field though increasing the number of layers in a hierarchical architecture. BEHRT's feedforward structure alleviates the exploding and vanishing gradient problems and capture information by considering the full sequence at the same time; a more efficient training through learning the data in parallel rather than in sequence (unlike the RNN).

The embedding layer in BEHRT, as shown in Figure~\ref{fig:BEHRT}, learns the evolution of one's EHR through a combination of four embeddings: disease, "position", age, and "segment". This combination enables BEHRT to define a representation that can capture one's EHR in as much detail as possible. Disease codes are of course important in informing the model of the future state of one's health. That is, there many common disease trajectories and multi-morbidity patterns~\cite{TheAcademyofMedicalSciences2018Multimorbidity:Research} that knowing one's past diseases can improve the accuracy of the prediction. Positional encodings are either trainable or pre-determined encodings for a position to determine relative position of words in a sequence of words.
Pre-determined encodings are used in this paper to avoid weak learning of positional embedding caused by imbalanced distribution of medical sequence length.
Telling the network of a disease's position, enables it to capture the positional interactions among diseases. Given the feed-forward architecture of our network, positional encodings plays a key role in filling the gap resulting from the lack of a recurrent structure that was the most common/successful approach for learning from sequences. For BEHRT, we followed the same position encoding rule proposed by~\cite{Vaswani2017}.

Age and segment are two embeddings that did not exist in the original BERT implementation for NLP and is unique to BEHRT; an attempt to empower it for dealing with the challenges we mentioned earlier. The segment embedding can be either A or B; the purpose of this is to use two trainable vectors to provide extra information for visit separation. Age, of course, is known to be a key risk factor in epidemiology. By embedding age and linking it to each visit/diagnosis, not only we provide the model with a sense of time (i.e., the time between events, as well as a universal notion of time for when things happened that is comparable across patients).

Through a unique combination of the four aforementioned embeddings, we not only provide the model with disease sequences, but also give it with a precise sense of timing of events, and data about the delivery of care. In other words, the model has the ability to learn from the diagnosis history, the age at which diagnoses happened and the pattern at which the patient was visited. All these, when combined, can provide a picture of one's health that traditionally we might have sought to paint through additional features extracted from EHR data. Of course, we do not advocate for not using the full richness of the EHR, however, the complexity of our architecture when paired with simple subset of EHR can still provide an accurate prediction of one's future health. Plus, BEHRT's flexible architecture enables the use of additional concepts, e.g., by simply adding a fifth or sixth (or more) embedding to the existing four.

\subsection{Pre-training BEHRT using MLM}
\label{mlm}
In EHR, just like language, it is intuitively reasonable to believe that a deep bidirectional model is more powerful than either a left-to-right model or the shallow concatenation of a left-to-right and a right-to-left model. Therefore, we pre-trained BEHRT using the same approach as the original BERT paper~\cite{Devlin2018}, using MLM. That is, we chose 15\% of the disease words at random, and modified them according to the following probabilities:
\begin{enumerate}
\item 80\% of the times replace them word with [MASK]
\item 10\% of the times replace them with a random disease word
\item 10\% of the times do nothing and keep them unchanged.
\end{enumerate}

Under this setting, BEHRT does not know which of the disease words are masked, so it stores a contextual representation of all of the disease words. Additionally, given the small prevalence of change (i.e., only for 15\% of all disease words) will not hamper the model's ability to understand the EHR language. Lastly, the replacement of the disease words acts as injected noise into the model; it will distract the model from learning the true left and right context, and instead forces the model to fight through the noise and continue learning the overall disease trajectories. The pre-training MLM task was evaluated using precision score~\cite{Powers2007}, % [RK] is this precision score, as in "threshold at p=0.5 and measure precision?" or average precision score, as you describe later?
which calculates the ratio of true positive over the number of predicted positive samples (precision calculated at a threshold of 0.5). The average is calculated over all labels and over all patients.

\subsection{Disease Prediction}
\label{pred}
In order to provide a comprehensive evaluation of BEHRT, we assess its learning for three predictive tasks: prediction of concepts in the next visit (T1), prediction of the occurrence of diseases in the next 6 months (T2), and prediction of the occurrence of diseases in the next 12 months (T3). In order to train our model and assess the goodness of its predictions across these tasks, we first randomly allocated the patients into three groups of train, validation and test (each containing 80\%, 10\% and 10\%, or patients, respectively). To define the training examples (i.e., input-output pairs) for T1, we randomly choose an index $j$ ($3<j<n_p$) for each patient and form $\xb_p=\{\vb_{p}^1, \ldots, \vb_{p}^j\}$ and $\yb_p=\wb_{j+1}$, as input and output, respectively, where $\wb_{j+1}$ is a multi-hot vector of length $G$, with 1 for diseases that exist in $\vb_{p}^{j+1}$. Note that, for each patient, we have only input-output pair.

For both T2 and T3, the formation of input and label are slightly modified. First, patients that do not have 6 or 12 months (for T2 and T3, respectively) worth of EHR (with or without a visit) after $\vb_{p}^4$ will not be included in these analyses. Second, $j$ is chosen randomly from $(3,n_*)$, where $n_*$ denotes the highest index after which there is 6 or 12 months (for T2 and T3, respectively) worth of EHR (with or without a visit). Lastly, $\yb_p=\wb_{6m}$ and $\yb_p=\wb_{12m}$ are multi-hot vectors of length $G$, with 1 for concepts/diseases that exist in the next 6 and 12 months, respectively. As a result of this final filtering of patients, we had 699K, 391K, and 342K patients for T1, T2, and T3, respectively.

We denote the model's prediction for patient $p$ in the aforementioned tasks as $\yb_p^*$, where the $i$th entry is the model's prediction of that person having $d_i$. The evaluation metrics we used to compare $\yb$s and $\yb^*$s, are area under the ROC curve (AUROC)~\cite{Fawcett2006} and average precision score (APS) \cite{Zhu2004}; the latter is a weighted mean of precision and recall numbers achieved at different thresholds. We calculated the APS and AUROC for each patient first, and then averaged the resulting APS and AUROC scores across all patients; this average is the key metric, when comparing BEHRT with other state-of-the-art architectures in the field. The methods we used here for APS and AUROC are described in~\cite{Fawcett2006, Zhu2004}.

\section{Results}

To begin with, we used Bayesian Optimisation~\cite{SnoekJasper;LarochelleHugo;Adams2017} to find the optimal hyperparameters for the MLM pre-training. The main hyperparameters here are the number of layers, the number of attention heads, hidden size, and intermediate size. Intermediate size is the size of the neural network layer titled "intermediate layer". % [RK] what is the last hyperparameter?
This process resulted in an optimal architecture with 6 layers, 12 attention heads, intermediate layer size of 512, and hidden size of 288; model's performance in the MLM task described in Section~\ref{mlm} was 0.6597 in precision score. Further details can be found in Appendix~\ref{app.hp}.

\begin{figure}[h]
    \centering
    \includegraphics[width=1\textwidth]{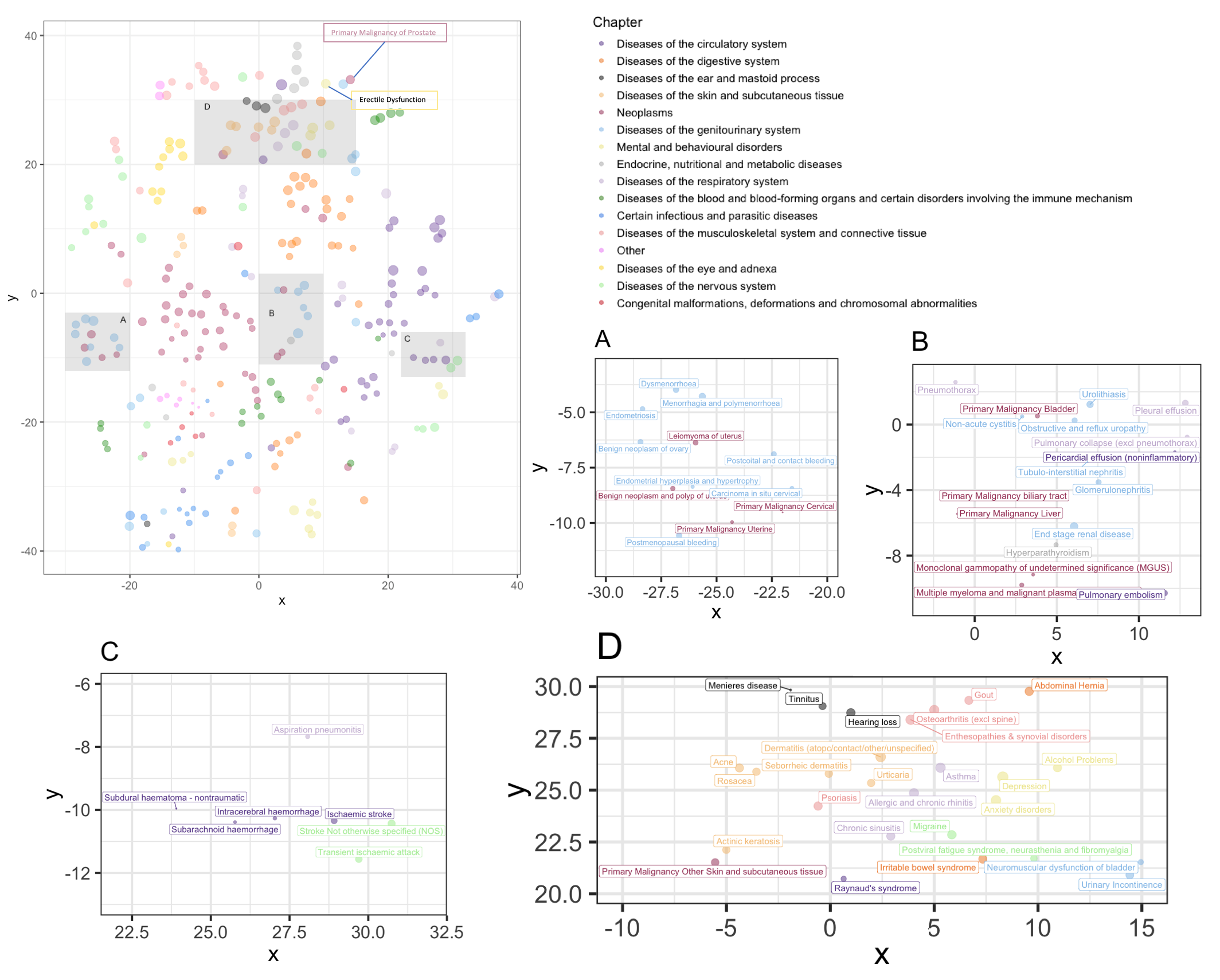}
    \caption{Disease Embedding Analysis. In this image we see a graph of disease embeddings projected in two dimensions where distance represents closeness of contextual association. Most associations are accepted by medical experts and maintain the gender-based divisions in illnesses, among other things. We zoom in and profile four clusters in this plot -- shown in subfigures A-D.} % [RK] colors need to improve, as they are opaque and hard to see (or, maybe a sign of ageing for me).
    \label{fig:embed2}
\end{figure}

In order for BEHRT's numerical processes to be applicable to EHR, we first need to map the non-numeric concepts such as diagnoses to a vector space (i.e., disease embedding). Therefore, we start the results by showing the performance of our pre-training process in embedding the diseases, where we mapped each of the $G$ diseases into a 288-dimensional vector. Note that, for evaluating an embedding technique -- even in NLP where the literature has a longer history and hence is more mature -- there is not a single gold standard metric~\cite{Wang}. In this study, our assessment is based on two techniques: visual investigation (i.e., in comparison with medical knowledge), and evaluation in a prediction task. For the former, we used t-SNE~\cite{Maaten2008} to reduce the dimensionality of the disease vectors to two -- results are shown in Figure~\ref{fig:embed2}. Based on the resulting patterns in lower dimension, we can see that diseases that are known to co-occur and/or belong to the same clinical groups, are grouped together.

A reassuring pattern that can be seen in Figure~\ref{fig:embed2} is the natural stratification of gender-specific diseases. For instance, diseases that are unique to women (e.g., Endometriosis, Dysmenorrhea, Menorrhagia, ...) are quite distant from those that are unique to men (e.g., Erectile Dysfunction, Primary Malignancy of Prostate, ...). Such patterns seem to suggest that our disease embedding built an understanding of the context in which diagnoses happen, and hence infer factors such as gender that it is not explicitly fed.

Furthermore, the colour in Figure~\ref{fig:embed2} represent the original Caliber disease chapters (see the legends in the main subplot). As can be seen, natural clusters are formed that in most cases consist of disease of the same chapter (i.e., the same colour). Some of these clusters, however, are correlated but not identical to these chapters; for instance, many Eye and Adnexa diseases are amongst nervous system diseases and many nervous system disease are also among many musculoskeletal diseases. Overall, this map can be seen as diseases' correspondence to each other based on 1.6 million people's EHR. Overall, it seems to be safe to say that this embedding seems to make sense and hence it passes the visual evaluation test.

\begin{figure}[h]
    \centering
    \includegraphics[width=1\textwidth]{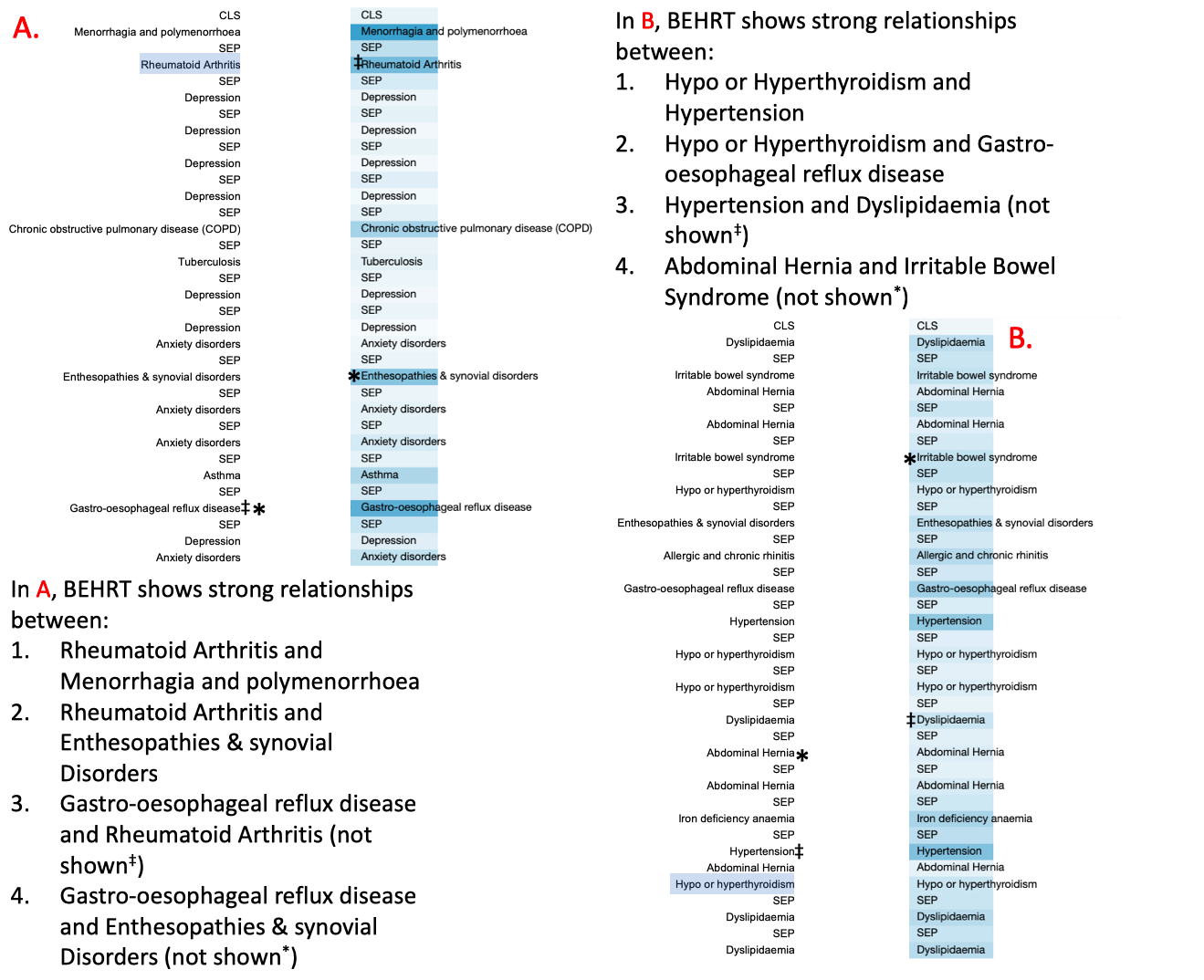}
    \caption{Disease Self-Attention Analysis. This figure shows the EHR history (shown chronologically, going downwards) of two patients A and B, each presented as two identical columns for the convenience of association analyses. The left side of the column represents the disease of interest and the right column indicates the corresponding associations to the highlighted disease on the left. The intensity of the blue on the right column represents the strength of the attention score -- the stronger the intensity, the stronger the association and hence the stronger the attention score. The attention scores are specifically retrieved from the attention component of the last layer of BEHRT network.}
    \label{fig:attention1}
\end{figure}

Another interesting property of BEHRT is its self-attention mechanism; it gives our model the ability to find the relationships among events that go beyond temporal/sequence adjacency. This self-attention mechanism is able to unearth deeper and more complex relationships between a disease in one visit and other surrounding diagnoses. We visualise this self-attention using the approach introduced in~\cite{Vig2019VisualizingModels}; we plot each medical history against itself, and we see how each disease relates to others around it in each patient. The results from a couple of example patients are shown in Figure~\ref{fig:attention1}. Note that, since BEHRT is bidirectional, the self-attention mechanism captures non-temporal/non-directional relationships among diseases.

For patient A for example, the self-attention mechanism has shown strong connections between Rheumatoid Arthritis and Enthesopathies and synovial disorders (far in the future of the patient). This is a great example of where attention can go beyond recent events and find long-range dependencies among diseases. Note that, as described earlier and illustrated in Figure~\ref{fig:BEHRT}, the sequence we model is a combination of four embedding (disease, plus age, segment and position) that go through layers of transformations to form a latent space abstraction. While in Figure~\ref{fig:attention1} we labeled the cells with disease names, a more precise labelling will be diseases in their context (e.g., at a given age).

BEHRT after the MLM pre-training can be considered a universal EHR feature extractor that with small additional training can be employed for a range of downstream tasks. In this work, the downstream task of choice is the multi-disease prediction problem that we described in Section~\ref{pred}. To train a predictor, we feed the output from BEHRT to a single feed-forward classifier layer and train it three separate time for each of the three tasks (T1-T3) described in Section~\ref{pred}. The evaluation of the model's performance is shown in Table~\ref{tab:performance}, which demonstrates BEHRT's superior predictive power compared to two of the most successful approaches in the literature (i.e., RETAIN~\cite{Choi2016a} and DeepR~\cite{Miotto2016DeepRecordsb}). We used Bayesian Optimisation to find the optimal hyperparameters for RETAIN and Deepr before assessing them in terms of APS and AUROC scores. More details on their hyperparameter search and optimisation can be found in Appendix\ref{app.hp}.

\begin{table}[h]
 \caption{Next Visit Prediction Task}
  \centering
\begin{tabular}{p{2.5cm}p{4cm}p{4cm}p{4cm}}
    \toprule
    Model Name & Next~Visit (APS~|~AUROC) & Next~6M (APS~|~AUROC) & Next~12M (APS~|~AUROC) \\
    \midrule
\bfseries    BEHRT & \bfseries 0.462~|~0.954 & \bfseries 0.525~|~0.958 & \bfseries 0.506~|~0.955 \\
    DeepR & 0.360~|~0.942 & 0.393~|~0.943 &  0.393~|~0.943\\
    RETAIN & 0.382~|~0.921 &  0.417~|~0.927 & 0.413~|~0.928\\
    \bottomrule
  \end{tabular}
  \label{tab:performance}
\end{table}

\begin{figure}[h]  % [RK] use A, B, C; also, might be worth adding similar figs for T1 and T3 in the Appendix. Also, do we have such results from DeepR and RETAIN?
    \includegraphics[width=1\textwidth]{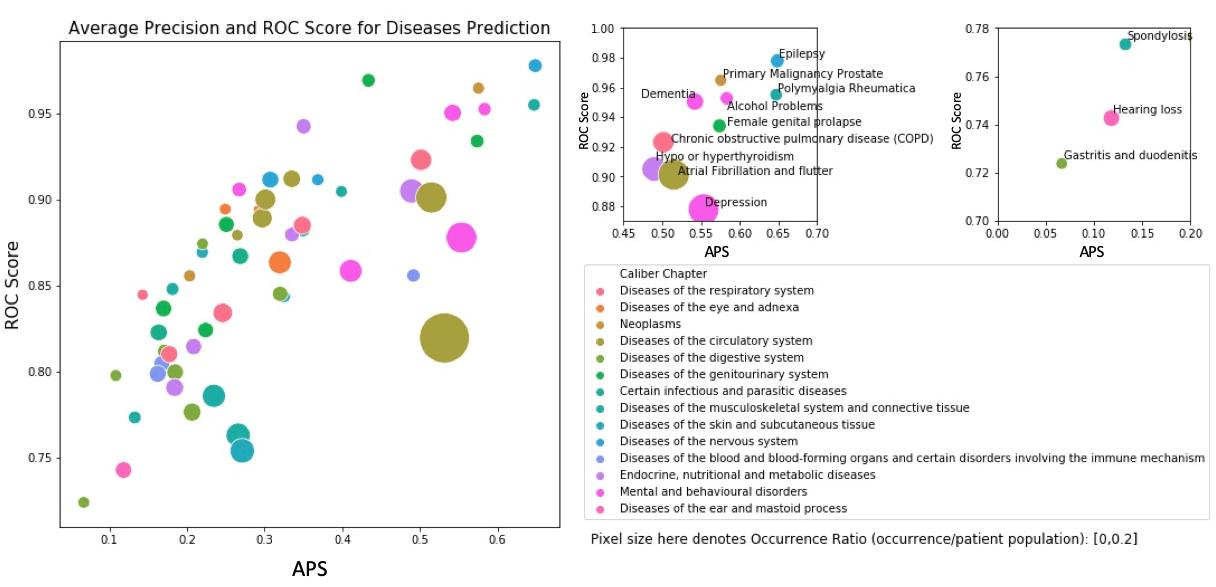}
    \caption{Disease-wise precision analysis. Each circle in these graphs represents a disease, who and color and size are their caliber chapter and prevalence, respectively. Also, in these plots, we show APS and AUROC on the x- and y-axis, respectively. Therefore, the further right and higher a disease, the better BEHRT's job at predicting its occurrence in the next 6 months. Subplot A, illustrated the full results, and subplots B and C illustrate the best and worth sections of the plot, in terms of BEHRT's performance.}
    \label{fig:prec}
\end{figure}

Besides comparing the APS, which provides an average view across all patients, all diseases and all thresholds, we are also interested in analysing the model's performance for predicting for each disease. To do so for a given disease $d_i$, we only considered the $i$th location in $\yb_p$ and $\yb_p^*$ vectors and calculated AUROC and APS scores, as well as occurrence ratio for comparison. The results for T2 (or, next 6-months prediction task) is shown in Figure \ref{fig:prec}. For visual convenience, we did not include rare diseases with prevalence of less than 1\% in our data. The result shows that the model is able to make predictions with relatively high precision and recall for diseases such as Epilepsy (0.016), Primary Malignancy Prostate (0.011), Polymyalgia Rheumatica (0.013), Hypo or hyperthyroidism (0.047) and Depression (0.0768). A numerical summary of this analysis can be found in be found in Appendix~\ref{app:dis_scores}. Furthermore, a comparison of the general APS/AUROC trends across the three models - BEHRT, RETAIN, and Deepr can be found in Appendix~\ref{app:comparison_scores}.

\section{Conclusions and Future Works}
\label{conclusions}
In this paper, we introduced a novel deep neural model for EHR called BEHRT, which can be pre-trained on a large dataset and then with small fine tuning result in a striking performance in a wide range of downstream tasks. We demonstrated this property of the model by training and testing it on CPRD - one of the largest linked primary care EHR systems. Based on our results, BEHRT outperformed all the best deep EHR models in the literature over a range of diseases (i.e., in a multi-label diagnoses prediction in near future), by $\sim$8\% (absolute improvement) in any given task.

BEHRT offers a flexible architecture that is capable of capturing more modalities of EHR data. In this paper, we designed and tested a BEHRT model that relied on 4 key embeddings - diseases, age, segment and position. Through this mix, the model will not only have the ability to learn about the past diseases and their relationships with one another (and hence their effect on the next likely disease), but also gain insights about the underlying generating process of EHR; we can refer to this as practice of care. In other words, the model will engineer features (i.e., complex representations) that are capable of capturing concepts such as "this patient had diseases X and Y at young ages, and suddenly, the frequency of visits increased and new diagnoses appeared, which can lead to high chance of the next disease being Z". In future works, one can add more to the four concepts we employed and bring medication, tests and interventions to the model with minimum architectural changes - only a vector addition in Figure~\ref{fig:BEHRT}.

%[comment] We have also laid out disease-wise Precision Score measurements in Supplementary Table \ref{tab: pre_full}. We see that many models train and test well on tailored and small datasets, but fail to capture the inherent complexities of patients and medical histories out in the real world. Since we have trained and tested on such a complex and diverse dataset as CPRD, we believe that BEHRT will be a good tool for diagnoses prediction in the real world as well.

Our primary objective in this study was to provide the field with an accurate predictive models for the prediction of next diseases. However, BEHRT provides multiple byproducts that can be useful on their own and/or as the foundation preprocessing for the future works. For instance, the disease embeddings resulting from BEHRT can provide a great insight to how various diseases are related to each other - it goes beyond simple disease co-occurrence and rather learns to score closeness of diseases based on their trajectories in a big population of patients. Furthermore, the disease correspondence that results from BEHRT's attention mechanism has been shown to be a useful tool for illustrating the disease trajectories for multi-morbid patients; not only it shows how diseases co-occur, but also it shows the influence of certain disease in the past on future diseases of interest. These correspondences are not strictly temporal but rather contextual. As a future work, we aim to provide these attention-visualisation tools to medical researchers to help them better understand the contextual meaning of a diagnoses in the midst of other diagnoses of patients. Through this tool, medical researchers can even craft medical history timelines based on certain diseases or patterns and in a way, query our BEHRT model and visualiser to perhaps uncover novel disease contexts.

For the future, we wish to make improvements to our model and use ensembles of BEHRTs and variations for better predictive power. Furthermore, we also plan to bring in more medical features such as treatment records and more demographic information (region, ethnicity, etc). Also, as discussed before, since Caliber consolidates specific codes into 301 disease codes and fails to map many other codes, we wish to look into more stable mappings that preserve the comprehensiveness of CPRD and reduce noise in our dataset. In addition to this, we wish to also embark on a deep dive of a single disease, perhaps heart failure or hypertension, to use BEHRT's diagnostic power for specific disease prediction.

\section{Acknowledgements}
This research was funded by the Oxford Martin School (OMS) and supported by the National Institute for Health Research (NIHR) Oxford Biomedical Research Centre (BRC). The views expressed are those of the authors and not necessarily those of the OMS, the UK National Health Service (NHS), the NIHR or the Department of Health and Social Care. This work uses data provided by patients and collected by the NHS as part of their care and support and would not have been possible without access to this data. The NIHR recognises and values the role of patient data, securely accessed and stored, both in underpinning and leading to improvements in research and care. We also thank Wayne Dorrington for his work in creating figures for this paper (Figure \ref{fig:cprd_introduction} and Figure \ref{fig:BEHRT}).

\newpage
\bibliographystyle{unsrt}

\bibliography{mainfile}

\begin{thebibliography}{10}

\bibitem{Ardila2019}
Diego Ardila, Atilla~P Kiraly, Sujeeth Bharadwaj, Bokyung Choi, Joshua~J
  Reicher, Lily Peng, Daniel Tse, Mozziyar Etemadi, Wenxing Ye, Greg Corrado,
  and David~P Naidich.
\newblock {End-to-end lung cancer screening with three-dimensional deep
  learning on low-dose chest computed tomography}.
\newblock {\em Nature Medicine}, 25(June), 2019.

\bibitem{Poplin2018}
Ryan Poplin, Avinash~V Varadarajan, Katy Blumer, Yun Liu, Michael~V McConnell,
  Greg~S Corrado, Lily Peng, and Dale~R Webster.
\newblock {Prediction of cardiovascular risk factors from retinal fundus
  photographs via deep learning}.
\newblock {\em Nature Biomedical Engineering}, 2(3):158--164, 2018.

\bibitem{Topol2019}
Eric~J Topol.
\newblock human and artificial intelligence.
\newblock {\em Nature Medicine}, 25(January), 2019.

\bibitem{Esteva2019AHealthcare.}
Andre Esteva, Alexandre Robicquet, Bharath Ramsundar, Volodymyr Kuleshov, Mark
  DePristo, Katherine Chou, Claire Cui, Greg Corrado, Sebastian Thrun, and Jeff
  Dean.
\newblock {A guide to deep learning in healthcare.}
\newblock {\em Nature medicine}, 25(1):24--29, 2019.

\bibitem{Sudlow2015UKAge}
Cathie Sudlow, John Gallacher, Naomi Allen, Valerie Beral, Paul Burton, John
  Danesh, Paul Downey, Paul Elliott, Jane Green, Martin Landray, Bette Liu,
  Paul Matthews, Giok Ong, Jill Pell, Alan Silman, Alan Young, Tim Sprosen, Tim
  Peakman, and Rory Collins.
\newblock {UK Biobank: An Open Access Resource for Identifying the Causes of a
  Wide Range of Complex Diseases of Middle and Old Age}.
\newblock {\em PLOS Medicine}, 12(3):e1001779, 2015.

\bibitem{Shickel2018DeepAnalysis}
Benjamin Shickel, Patrick Tighe, Azra Bihorac, and Parisa Rashidi.
\newblock {Deep EHR: A Survey of Recent Advances in Deep Learning Techniques
  for Electronic Health Record (EHR) Analysis}.
\newblock {\em IEEE Journal of Biomedical and Health Informatics},
  22(5):1589--1604, 2018.

\bibitem{Meeting2018}
O~N C~Annual Meeting.
\newblock {Electronic Public Health Reporting}.
\newblock {\em None}, 2018.
\newblock Available at:
  \url{https://www.healthit.gov/sites/default/files/2018-12/ElectronicPublicHealthReporting.pdf}.

\bibitem{Parasrampuria2019Hospitals2015-2017}
Sonal Parasrampuria and Jawanna Henry.
\newblock {Hospitals' Use of Electronic Health Records Data, 2015-2017}.
\newblock {\em ONC Data Brief}, No. 46, 2019.

\bibitem{Rahimian2018PredictingRecords}
Fatemeh Rahimian, Gholamreza Salimi-Khorshidi, Amir~H Payberah, Jenny Tran,
  Roberto {Ayala Solares}, Francesca Raimondi, Milad Nazarzadeh, Dexter Canoy,
  and Kazem Rahimi.
\newblock {Predicting the risk of emergency admission with machine learning:
  Development and validation using linked electronic health records}.
\newblock {\em PLoS Medicine}, 15(11):1--18, 2018.

\bibitem{Liang2014}
Znaonui Liang, Gang Zhang, Jimmy~Xiangji Huang, and Qmming~Vivian Hu.
\newblock {Deep learning for healthcare decision making with EMRs}.
\newblock {\em Proceedings - 2014 IEEE International Conference on
  Bioinformatics and Biomedicine, IEEE BIBM 2014}, pages 556--559, 2014.

\bibitem{Tran2015LearningeNRBM}
Truyen Tran, Tu~Dinh Nguyen, Dinh Phung, and Svetha Venkatesh.
\newblock {Learning vector representation of medical objects via EMR-driven
  nonnegative restricted Boltzmann machines (eNRBM)}.
\newblock {\em Journal of Biomedical Informatics}, 2015.

\bibitem{Miotto2016DeepRecordsb}
Riccardo Miotto, Li~Li, Brian~A Kidd, and Joel~T Dudley.
\newblock {Deep Patient: An Unsupervised Representation to Predict the Future
  of Patients from the Electronic Health Records}.
\newblock {\em Scientific Reports}, 6(May):1--10, 2016.

\bibitem{Nguyen2017}
Phuoc Nguyen, Truyen Tran, Nilmini Wickramasinghe, and Svetha Venkatesh.
\newblock {Deepr: A Convolutional Net for Medical Records}.
\newblock {\em IEEE Journal of Biomedical and Health Informatics},
  21(1):22--30, may 2017.

\bibitem{Choi2016DoctorNetworks.}
Edward Choi, Mohammad~Taha Bahadori, Andy Schuetz, Walter~F Stewart, and Jimeng
  Sun.
\newblock {Doctor AI: Predicting Clinical Events via Recurrent Neural
  Networks.}
\newblock {\em JMLR workshop and conference proceedings}, 56:301--318, 2016.

\bibitem{Pham2016}
Trang Pham, Truyen Tran, Dinh Phung, and Svetha Venkatesh.
\newblock {DeepCare: A deep dynamic memory model for predictive medicine}.
\newblock {\em Lecture Notes in Computer Science (including subseries Lecture
  Notes in Artificial Intelligence and Lecture Notes in Bioinformatics)}, 9652
  LNAI(i):30--41, 2016.

\bibitem{Choi2016a}
Edward Choi, Mohammad~Taha Bahadori, Joshua~A. Kulas, Andy Schuetz, Walter~F.
  Stewart, and Jimeng Sun.
\newblock {RETAIN: An Interpretable Predictive Model for Healthcare using
  Reverse Time Attention Mechanism}.
\newblock {\em arxiv}, 2016.

\bibitem{Solares}
Jose Roberto~Ayala Solares, Francesca Elisa~Diletta Raimondi, Yajie Zhu,
  Fatemeh Rahimian, Dexter Canoy, Jenny Tran, Ana Catarina~Pinho Gomes, Amir
  Payberah, Mariagrazia Zottoli, Milad Nazarzadeh, Nathalie Conrad, Kazem
  Rahimi, and Gholamreza Salimi-Khorshidi.
\newblock {Deep Learning for Electronic Health Records: A Comparative Review of
  Multiple Deep Neural Architectures}.
\newblock {\em preprint}, page~57, 2019.

\bibitem{Devlin2018}
Jacob Devlin, Ming-Wei Chang, Kenton Lee, and Kristina Toutanova.
\newblock {BERT: Pre-training of Deep Bidirectional Transformers for Language
  Understanding}.
\newblock {\em arxiv}, 2018.

\bibitem{Herrett2015}
Emily Herrett, Arlene~M Gallagher, Krishnan Bhaskaran, Harriet Forbes, Rohini
  Mathur, Tjeerd~Van Staa, and Liam Smeeth.
\newblock {Data Resource Profile: Clinical Practice Research Datalink (CPRD)}.
\newblock {\em International Journal of Epidemiology}, 44(3):827--836, 2015.

\bibitem{WALLEY19971097}
T~Walley and A~Mantgani.
\newblock The uk general practice research database.
\newblock {\em The Lancet}, 350(9084):1097 -- 1099, 1997.

\bibitem{Emdin2015UsualAdults}
Connor~A Emdin, Simon~G Anderson, Thomas Callender, Nathalie Conrad, Gholamreza
  Salimi-Khorshidi, Hamid Mohseni, Mark Woodward, and Kazem Rahimi.
\newblock {Usual blood pressure, peripheral arterial disease, and vascular
  risk: Cohort study of 4.2 million adults}.
\newblock {\em BMJ (Online)}, 2015.

\bibitem{Emdin2017}
Connor~A. Emdin, Simon~G. Anderson, Gholamreza Salimi-Khorshidi, Mark Woodward,
  Stephen MacMahon, Terrence Dwyer, and Kazem Rahimi.
\newblock {Usual blood pressure, atrial fibrillation and vascular risk:
  Evidence from 4.3 million adults}.
\newblock {\em International Journal of Epidemiology}, 2017.

\bibitem{Lee2002}
F.~Lee, H.~R.S. Patel, and M.~Emberton.
\newblock {The 'top 10' urological procedures: A study of hospital episodes
  statistics 1998-99}.
\newblock {\em BJU International}, 2002.

\bibitem{Mohseni2017}
Hamid Mohseni, Amit Kiran, Reza Khorshidi, and Kazem Rahimi.
\newblock {Influenza vaccination and risk of hospitalization in patients with
  heart failure: A self-controlled case series study}.
\newblock {\em European Heart Journal}, 2017.

\bibitem{NHS}
NHS.
\newblock {Read Codes}.
\newblock Available at:
  \url{https://digital.nhs.uk/services/terminology-and-classifications/read-codes}.

\bibitem{WHO}
WHO.
\newblock {ICD-10 online versions}.
\newblock Available at \url{https://icd.who.int/browse10/2016/e}.

\bibitem{Kuan2019}
Valerie Kuan, Spiros Denaxas, Arturo Gonzalez-izquierdo, Kenan Direk, Osman
  Bhatti, Shanaz Husain, Shailen Sutaria, Melanie Hingorani, Dorothea Nitsch,
  Constantinos~A Parisinos, R~Thomas Lumbers, Rohini Mathur, Reecha Sofat,
  Juan~P Casas, Ian C~K Wong, and Harry Hemingway.
\newblock {Articles A chronological map of 308 physical and mental health
  conditions from 4 million individuals in the English National Health
  Service}.
\newblock {\em The Lancet Digital Health}, 1(2):e63--e77, 2019.

\bibitem{Cho2013}
Kyunghyun Cho.
\newblock {Learning Phrase Representations using RNN Encoder–Decoder for
  Statistical Machine Translation}.
\newblock {\em arxiv}, 2013.

\bibitem{Vaswani2017}
Ashish Vaswani, Noam Shazeer, Niki Parmar, Jakob Uszkoreit, Llion Jones,
  Aidan~N Gomez, Lukasz Kaiser, and Illia Polosukhin.
\newblock {Attention Is All You Need}.
\newblock {\em arXiv:1706.03762 [cs]}, apr 2017.

\bibitem{Pascanu2012OnNetworks}
Razvan Pascanu, Tomas Mikolov, and Yoshua Bengio.
\newblock {On the difficulty of training Recurrent Neural Networks}.
\newblock {\em arxiv}, 2012.

\bibitem{TheAcademyofMedicalSciences2018Multimorbidity:Research}
{The Academy of Medical Sciences}.
\newblock {Multimorbidity: a priority for global health research}.
\newblock {\em The Academy of Medical Sciences}, pages 1--127, 2018.

\bibitem{Powers2007}
David M~W Powers.
\newblock {Evaluation : From Precision , Recall and F-Factor to ROC ,
  Informedness , Markedness {\&} Correlation}.
\newblock {\em arxiv}, 2007.

\bibitem{Fawcett2006}
Tom Fawcett.
\newblock {An introduction to ROC analysis}.
\newblock {\em Pattern Recognition Letters}, 2006.

\bibitem{Zhu2004}
Mu~Zhu.
\newblock {Recall, precision and average precision}.
\newblock {\em Department of Statistics and Actuarial Science, {\ldots}}, 2004.

\bibitem{SnoekJasper;LarochelleHugo;Adams2017}
Ryan {Snoek, Jasper; Larochelle, Hugo; Adams}.
\newblock {Practical Bayesian Optimization of Machine Learning Algorithms}.
\newblock {\em NIPS}, 2(12):e540, 2017.

\bibitem{Wang}
Bin Wang, Student Member, Angela Wang, Fenxiao Chen, Student Member, Yuncheng
  Wang, and C~Jay Kuo.
\newblock {Evaluating Word Embedding Models : Methods and Experimental
  Results}.
\newblock {\em arxiv}, pages 1--13, 2019.

\bibitem{Maaten2008}
Laurens Van~Der Maaten and Geoffrey Hinton.
\newblock {Visualizing Data using t-SNE}.
\newblock {\em JMLR}, 9:2579--2605, 2008.

\bibitem{Vig2019VisualizingModels}
Jesse Vig.
\newblock {Visualizing Attention in Transformer-Based Language Representation
  Models}.
\newblock {\em arxiv}, pages 2--7, 2019.

\end{thebibliography}

\newpage

\appendix

% \section{BEHRT, Transformer and BERT}

\section{Hyperparameter Tuning} % [RK] While it might appear as unnecessary, it's good to add a line or two of description. it is common practice to set the scene up (it must read on its own).
We show the hyperparameter tuning results here in the following section. In Table \ref{tab:tableBayes}, we show the results of the MLM training hyperparameter tuning process. We performed Bayesian Optimization to retrieve optimal parameters for the model.\\
In the following sections, we also perform hyperparameter searches for the Deepr (Table \ref{tab:tableBayesDeepr}) and RETAIN (Table \ref{tab:tableBayesRETAIN}) models to ensure proper comparison of model performance between BEHRT and the aforementioned models.

\label{app.hp}

\begin{table}[h]
\renewrobustcmd{\bfseries}{\fontseries{b}\selectfont}
\renewrobustcmd{\boldmath}{}
 \caption{MLM Hyperparameter Tuning}
  \centering
\begin{tabular}{p{1cm}p{2cm}p{1cm}p{2.5cm}p{2.5cm}p{3cm}}
    \toprule
    \cmidrule(r){1-2}
   Iteration     & Hidden Size    & Layers & Attention Heads &  Intermediate Size      &  Precision \\
    \midrule
   1 &  216 & 3 &6 & 256 & 0.6191     \\
   2 &288& 9& 12& 512 &  0.6399 \\
    3 & 216& 3& 12& 512 & 0.6175 \\
   4 & 432& 3& 18& 512 &0.6397\\
   5 & 288& 6& 6& 784 & 0.6380\\
 6 &   216& 6& 18& 512& 0.6262  \\
  7 & 288& 3& 18& 512 & 0.6292\\
  8 & 432& 3& 6& 784 & 0.6426\\
 9 &  288& 6& 12& 512 & 0.6356 \\
 10 &  288& 3& 12& 256 & 0.6283   \\
11 &432& 9& 18& 512& 0.6466   \\
12 &576& 9& 6& 1024& 0.6538  \\
 13 &432& 3& 18& 1024 & 0.6411 \\
14 & 432& 9& 6& 1024  & 0.6508 \\
15 & 576& 6& 6& 256 & 0.6503   \\
16 & 576& 6& 12& 256& 0.6510  \\
17 & 360& 9& 18& 512 & 0.6404   \\
18 & 576& 9& 6& 512 & 0.6513  \\
19 & 288& 6& 6& 512 & 0.6363  \\
20 & 288& 3& 6& 512 & 0.6297 \\
\bfseries21 & \bfseries288& \bfseries6& \bfseries12& \bfseries512 & \bfseries0.6597  \\
22 & 576& 3& 12& 512 & 0.6487   \\
23 & 360& 6& 6& 784  & 0.6412  \\
24 & 432& 9& 6& 512 & 0.6497 \\
25 & 360& 6& 12& 512 & 0.6423   \\
    \bottomrule
  \end{tabular}
  \label{tab:tableBayes}
\end{table}

\begin{table}[p]
 \caption{Deepr Best Model - Subsequent Visit Prediction Task}
  \centering
\begin{tabular}{p{1cm}p{1cm}p{1cm}p{1cm}p{1cm}p{1cm}p{1cm}p{1cm}p{1cm}p{1cm}p{3cm}}
    \toprule
    \cmidrule(r){1-2}
Iteration & Filters     & Kernel Size & FC I & FC II & FC III & Dropout I&  Dropout  II &  Dropout  III &Learning Rate       & Average Precision \\
    \midrule
1&37& 7& 10& 46& 28& 0.4139& 0.4997& 0.3718& 0.0004 &0.2599 \\
2&24& 5& 28& 19& 13& 0.4936& 0.1722& 0.4643& 0.0062 &0.2319
\\
3&17& 7& 16& 48& 40& 0.2250& 0.3945& 0.3264& 0.0310 &0.1815 \\
4&33& 7& 6& 10& 25& 0.1602& 0.4476& 0.3544& 0.0026&0.2640\\
5&32& 7& 4& 30& 32& 0.3714& 0.3382& 0.3573& 0.0353 &0.2005\\
6&13& 4& 50& 17& 47& 0.3424& 0.1830& 0.3056& 0.0008 &0.3274\\
7&12& 3& 3& 6& 43& 0.4201& 0.2387& 0.3884& 0.0123&0.2112 \\
8&30& 7& 16& 26& 41& 0.2055& 0.3343& 0.1541& 0.0011&0.3256\\
9&35& 5& 50& 24& 25& 0.1567& 0.4056& 0.1329& 0.0004&0.3433\\
10&41& 7& 9& 35& 19& 0.4885& 0.3548& 0.3080& 0.0004&0.2343\\
11&4& 4& 33& 12& 39& 0.1811& 0.1251& 0.1066& 0.0010&0.3051\\
12&48& 4& 36& 45& 37& 0.2143& 0.3486& 0.1222& 0.0015&0.3504\\
13&36& 3& 39& 10& 48& 0.1992& 0.4164& 0.1183& 0.0050&0.3291\\
14&45& 4& 44& 40& 26& 0.2329& 0.2900& 0.1280& 0.0903&0.0042\\
15&40& 4& 35& 48& 11& 0.1600& 0.4704& 0.1211& 0.0019 &0.3125\\
\bfseries16&\bfseries49& \bfseries3& \bfseries47& \bfseries41& \bfseries40& \bfseries0.1002 & \bfseries0.2541& \bfseries0.1284& \bfseries0.0019 &\bfseries0.3588\\
17&47& 3& 50& 47& 12& 0.2519& 0.2125& 0.1668& 0.0005 &0.2988\\
18&47& 3& 37& 35& 50& 0.1059& 0.4225& 0.1120& 0.0034&0.3487\\
19&47& 4& 48& 34& 39& 0.2177& 0.3607& 0.1301& 0.0016&0.3567\\
20&46& 3& 46& 45& 43& 0.2007& 0.1609& 0.1196& 0.0044&0.3356\\
    \bottomrule
  \end{tabular}
  \label{tab:tableBayesDeepr}
\end{table}

\begin{table}[p]
 \caption{RETAIN Best Model - Subsequent Visit Prediction Task}
  \centering
\begin{tabular}{p{1cm}p{2.3cm}p{2.3cm}p{2cm}p{2cm}p{0.7cm}p{2.5cm}}
    \toprule
    \cmidrule(r){1-2}
Iteration & Embedding Size & Recurrent Size & Dropout Embedding & Dropout Context & L2 & Average Precision \\
    \midrule
1& 142& 90& 0.3846& 0.0224& 0.0891 & 0.1822 \\
2&124& 43& 0.3922& 0.0382& 0.0003 &0.1815 \\
3&173& 90& 0.3929& 0.2238& 0.0014 &0.3479 \\
4&145& 91& 0.4117& 0.0404& 0.0102&0.2049\\
5&153& 92& 0.4569& 0.0642& 0.0116&0.2049\\
6&120& 37& 0.4335& 0.4017& 0.0728 &0.1815\\
7&180& 102& 0.3567& 0.3307& 0.0039&0.2469 \\
\bfseries8&\bfseries195&\bfseries 38& \bfseries0.3805& \bfseries0.2711& \bfseries0.0010& \bfseries0.3740\\
9&165& 119& 0.4969& 0.1964& 0.0047& 0.2292\\
10&174& 92& 0.3862& 0.2078& 0.0019& 0.3329\\
11&145& 110& 0.3928& 0.1926& 0.0891& 0.1813\\
12&195& 83& 0.4418& 0.4787& 0.0011& 0.3543\\
13&187& 110& 0.3456& 0.2083& 0.0123& 0.2122\\
14&144& 80& 0.3717& 0.0932& 0.0032&0.3038\\
15&193& 68& 0.4528& 0.3950& 0.0022&0.3280\\
16&198& 45& 0.4344& 0.3324& 0.0442&0.1828\\
17&145& 64& 0.4213& 0.1950& 0.0626&0.1813\\
18&171& 116& 0.4166& 0.4950&0.0028&0.3197\\
19&186& 38& 0.4062& 0.4916& 0.0011&0.3309\\
20&136& 54& 0.3503& 0.1678& 0.0067&0.2162\\
    \bottomrule
  \end{tabular}
  \label{tab:tableBayesRETAIN}
\end{table}
\newpage

\section{Disease-wise Model Performance} % [RK] While it might appear as unnecessary, it's good to add a line or two of description. it is common practice to set the scene up (it must read on its own).
Here we show Disease-wide BEHRT performance in terms of AUROC and APS. We have displayed codes with an occurrence ratio of 0.01 at the least. And detailed below is the description of the Caliber code and the chapter along with the APS/AUROC.\\
\label{app:dis_scores}
\begin{longtable}[h]{|p{1cm}|p{1.2cm}|p{1.2cm}|p{4.5cm}|p{1.5cm}|p{4.5cm}|}
\caption{Diseases Prediction Performance }\\
 \hline
Caliber &  APS &  AUROC & Description & Ratio & Caliber Chapter \\
 \hline
 \endhead
  92 &  0.066765 &  0.723828 &  Gastritis and duodenitis &  0.011198 &  Diseases of the digestive system \\
  66 &  0.108185 &  0.797633 &  Diaphragmatic hernia &  0.011490 &  Diseases of the digestive system \\
  100 &  0.118093 &  0.742646 &  Hearing loss &  0.021964 &  Diseases of the ear and mastoid process \\
  273 &  0.132567 &  0.773249 &  Spondylosis &  0.013459 &  Diseases of the musculoskeletal system and connective tissue \\
  189 &  0.142594 &  0.844596 &  Pleural effusion &  0.010229 &  Diseases of the respiratory system \\
  172 &  0.162186 &  0.798654 &  Other anaemias &  0.023303 &  Diseases of the blood and blood-forming organs and certain disorders involving the immune mechanism \\
  29 &  0.163355 &  0.822707 &  Bacterial Diseases (excl TB) &  0.023979 &  Certain infectious and parasitic diseases   \\
  130 &  0.167457 &  0.804545 &  Iron deficiency anaemia &  0.020780 &  Diseases of the blood and blood-forming organs and certain disorders involving the immune mechanism \\
  295 &  0.169674 &  0.836610 &  Urinary Tract Infections &  0.022534 &  Diseases of the genitourinary system \\
  69 &  0.170852 &  0.811759 &  Diverticular disease of intestine (acute and chronic) &  0.015966 &  Diseases of the digestive system \\
  8 &  0.177170 &  0.810068 &  Allergic and chronic rhinitis &  0.023241 &  Diseases of the respiratory system \\
  170 &  0.181182 &  0.847938 &  Osteoporosis &  0.013982 &  Diseases of the musculoskeletal system and connective tissue \\
  71 &  0.184000 &  0.790655 &  Dyslipidaemia &  0.026010 &  Endocrine, nutritional and metabolic diseases \\
  168 &  0.184629 &  0.799592 &  Oesophagitis and oesophageal ulcer &  0.022426 &  Diseases of the digestive system \\
  217 &  0.203218 &  0.855593 &  Primary Malignancy Other Skin and subcutaneous tissue &  0.012013 &  Neoplasms \\
  93 &  0.206455 &  0.776367 &  Gastro-oesophageal reflux disease &  0.026271 &  Diseases of the digestive system \\
  290 &  0.208524 &  0.814481 &  Type 1 Diabetes Mellitus, Type 2 Diabetes Mellitus, and Diabetes Mellitus – other or not specified &  0.021226 &  Endocrine, nutritional and metabolic diseases \\
  3 &  0.219637 &  0.869366 &  Actinic keratosis &  0.012490 &  Diseases of the skin and subcutaneous tissue \\
  131 &  0.220017 &  0.874319 &  Irritable bowel syndrome &  0.011182 &  Diseases of the digestive system \\
  294 &  0.223863 &  0.824114 &  Urinary Incontinence &  0.020057 &  Diseases of the genitourinary system \\
  169 &  0.234444 &  0.785766 &  Osteoarthritis (excl spine) &  0.043714 &  Diseases of the musculoskeletal system and connective tissue \\
  176 &  0.245906 &  0.834062 &  Other or unspecified infectious organisms &  0.030963 &  Diseases of the respiratory system \\
  95 &  0.249208 &  0.894444 &  Glaucoma &  0.011367 &  Diseases of the eye and adnexa \\
  109 &  0.250573 &  0.885547 &  Hyperplasia of prostate &  0.020842 &  Diseases of the genitourinary system \\
  184 &  0.264687 &  0.879325 &  Peripheral arterial disease &  0.010951 &  Diseases of the circulatory system \\
  79 &  0.265863 &  0.762939 &  Enthesopathies \& synovial disorders &  0.047036 &  Diseases of the musculoskeletal system and connective tissue \\
  81 &  0.267187 &  0.905812 &  Erectile dysfunction &  0.017873 &  Mental and behavioural disorders \\
  140 &  0.268504 &  0.867094 &  Lower Respiratory Tract Infections &  0.023518 &  Certain infectious and parasitic diseases   \\
  63 &  0.271027 &  0.753816 &  Dermatitis (atopc/contact/other/unspecified) &  0.049051 &  Diseases of the skin and subcutaneous tissue \\
  142 &  0.292598 &  0.893802 &  Macular degeneration &  0.010752 &  Diseases of the eye and adnexa \\
  274 &  0.296798 &  0.889236 &  Stable Angina &  0.032039 &  Diseases of the circulatory system \\
  57 &  0.301041 &  0.900088 &  Coronary heart disease not otherwise specified &  0.035177 &  Diseases of the circulatory system \\
  275 &  0.307238 &  0.911618 &  Stroke Not otherwise specified (NOS) &  0.023395 &  Diseases of the nervous system \\
  45 &  0.319433 &  0.863447 &  Cataract &  0.042099 &  Diseases of the eye and adnexa \\
  1 &  0.319972 &  0.845171 &  Abdominal Hernia &  0.019180 &  Diseases of the digestive system \\
  44 &  0.325143 &  0.843480 &  Carpal tunnel syndrome &  0.012013 &  Diseases of the nervous system \\
  101 &  0.334902 &  0.912117 &  Heart failure &  0.024918 &  Diseases of the circulatory system \\
  164 &  0.335131 &  0.879670 &  Obesity &  0.017442 &  Endocrine, nutritional and metabolic diseases \\
  22 &  0.348523 &  0.885055 &  Asthma &  0.026133 &  Diseases of the respiratory system \\
  97 &  0.349361 &  0.882694 &  Gout &  0.018058 &  Diseases of the musculoskeletal system and connective tissue \\
  65 &  0.350132 &  0.942604 &  Diabetic ophthalmic complications &  0.018919 &  Endocrine, nutritional and metabolic diseases \\
  147 &  0.368465 &  0.911541 &  Migraine &  0.012028 &  Diseases of the nervous system \\
  229 &  0.398842 &  0.904686 &  Psoriasis &  0.011751 &  Diseases of the musculoskeletal system and connective tissue \\
  17 &  0.410899 &  0.858498 &  Anxiety disorders &  0.041914 &  Mental and behavioural disorders \\
  146 &  0.433645 &  0.969406 &  Menorrhagia and polymenorrhoea &  0.015504 &  Diseases of the genitourinary system \\
  113 &  0.489456 &  0.905032 &  Hypo or hyperthyroidism &  0.047897 &  Endocrine, nutritional and metabolic diseases \\
  302 &  0.491672 &  0.855823 &  Vitamin B12 deficiency anaemia &  0.014489 &  Diseases of the blood and blood-forming organs and certain disorders involving the immune mechanism \\
  51 &  0.501496 &  0.923082 &  Chronic obstructive pulmonary disease (COPD) &  0.036869 &  Diseases of the respiratory system \\
  23 &  0.514881 &  0.901268 &  Atrial Fibrillation and flutter &  0.077629 &  Diseases of the circulatory system \\
  110 &  0.531597 &  0.819527 &  Hypertension &  0.200618 &  Diseases of the circulatory system \\
  61 &  0.542223 &  0.950442 &  Dementia &  0.024656 &  Mental and behavioural disorders \\
  62 &  0.553561 &  0.877904 &  Depression &  0.076876 &  Mental and behavioural disorders \\
  85 &  0.573880 &  0.934049 &  Female genital prolapse &  0.015781 &  Diseases of the genitourinary system \\
  220 &  0.575574 &  0.964776 &  Primary Malignancy Prostate &  0.011844 &  Neoplasms \\
  6 &  0.583305 &  0.952656 &  Alcohol Problems &  0.014535 &  Mental and behavioural disorders \\
  194 &  0.647243 &  0.955062 &  Polymyalgia Rheumatica &  0.013213 &  Diseases of the musculoskeletal system and connective tissue \\
  80 &  0.648763 &  0.977907 &  Epilepsy &  0.016104 &  Diseases of the nervous system \\
  \hline
\end{longtable}

\section{Comparison of APS/AUROC across Models} \label{app:comparison_scores}
\begin{figure}[h]  % [RK] use A, B, C; also, might be worth adding similar figs for T1 and T3 in the Appendix. Also, do we have such results from DeepR and RETAIN?
    \includegraphics[width=1\textwidth]{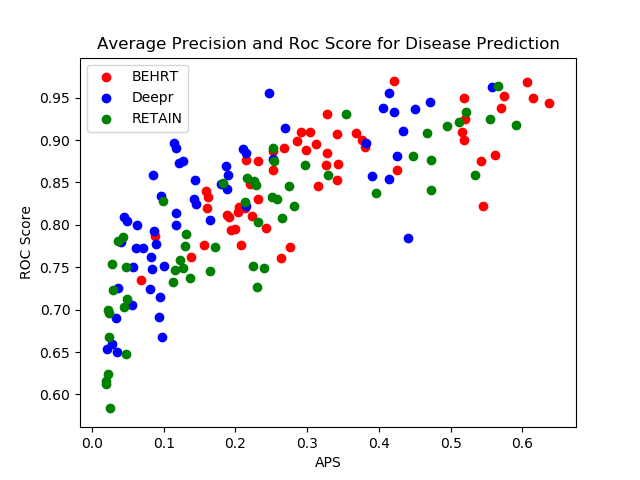}
    \caption{Disease-wise precision comparison for BEHRT, Deepr and RETAIN, all models trained on same dataset and trained on the same task (next 6 months), each point represents the precision/AUROC of one disease on corresponding model.}
    \label{fig:prec_compare}
\end{figure}

In Figure~\ref{fig:prec_compare}, we see the general trend of the three models. BEHRT's predictions remain in the upper right quadrant of the graph (for the most part) denoting higher APS/AUROC than the other two models.

\end{document}